\documentclass[runningheads]{llncs}

\usepackage[T1]{fontenc}
\usepackage[utf8]{inputenc}
\usepackage{graphicx}
\usepackage{esvect}
\usepackage{makecell}
\usepackage{tabularx}
\usepackage{array}
\usepackage{comment}
\usepackage{url}
\usepackage{amsmath}
\usepackage{booktabs}

\graphicspath{{LaTeX2e+Proceedings+Templates+download/}}

\newcolumntype{Y}{>{\raggedright\arraybackslash}X}
\begin{document}

\title{TDSal: A Task-Based Top-Down Saliency Prediction Model}
\titlerunning{TDSal: Task-Based Top-Down Saliency Prediction Model}

\author{Can Mizrakli\inst{1, 2}\orcidID{0009-0000-1990-3996} \and
Tolga K. Capin\inst{2}\orcidID{0000-0002-7843-6336}}
\authorrunning{C. Mizrakli and T. K. Capin}

\institute{Karlsruhe Institute of Technology, Germany\\
\and
TED University, Türkiye\\
\email{can.mizrakli@student.kit.edu, tolga.capin@tedu.edu.tr}}

\maketitle

\begin{abstract}
Visual saliency aims to predict the regions of an image most likely to attract human visual attention. While most saliency models assume free-viewing conditions, human attention is often shaped by explicit task goals. In this work, we address task-driven saliency prediction by proposing a model that conditions visual attention on natural-language task descriptions. The model produces task-dependent saliency maps that reflect how attention shifts under different viewing intents. Through quantitative and qualitative analysis, we show that incorporating explicit task semantics enables more faithful modeling of goal-directed visual attention.

\end{abstract}

\section{Introduction}
\label{sec:intro}
Visual saliency aims to predict where humans attend in an image, often modeled as a spatial probability map of fixation likelihood. Classical bottom-up approaches rely on low-level cues such as contrast, color, and orientation under free-viewing assumptions~\cite{itti2002model,itti2001computational}. However, in many real-world settings attention is goal-directed, and task demands can substantially shift where observers look.

Task-conditioned saliency prediction is becoming increasingly important in domains such as autonomous driving, human–robot interaction, and assistive vision. However, existing approaches often fall short in two critical aspects: (1) they do not explicitly incorporate high-level task semantics into the saliency generation process, and (2) they lack a lightweight mechanism for combining such semantic cues with spatial visual features. Approaches based on complex transformer or diffusion architectures~\cite{liu2021visual,zhang2024tdiffsal} can suffer from reduced spatial resolution and high computational cost, whereas methods without explicit task conditioning may fail to capture goal-directed human attention patterns.\footnote{Supplementary material associated with this work is publicly available through the project repository: https://github.com/canmizrakli/TDSal-2026}

To address these limitations, we structure our study around the following research questions:

\begin{itemize}
    \item \textbf{RQ1:} How can textual task definitions, encoded through sentence-level embeddings, be fused with spatial visual features to guide attention maps?
    \item \textbf{RQ2:} Can explicit vision--language conditioning support fixation alignment in task-driven saliency prediction?
    \item \textbf{RQ3:} 
    How can a pretrained object-detection backbone be leveraged to extract spatially rich visual features that support task-oriented saliency modeling?
\end{itemize}

\begin{figure}[t]
    \centering
    \includegraphics[width=\textwidth]{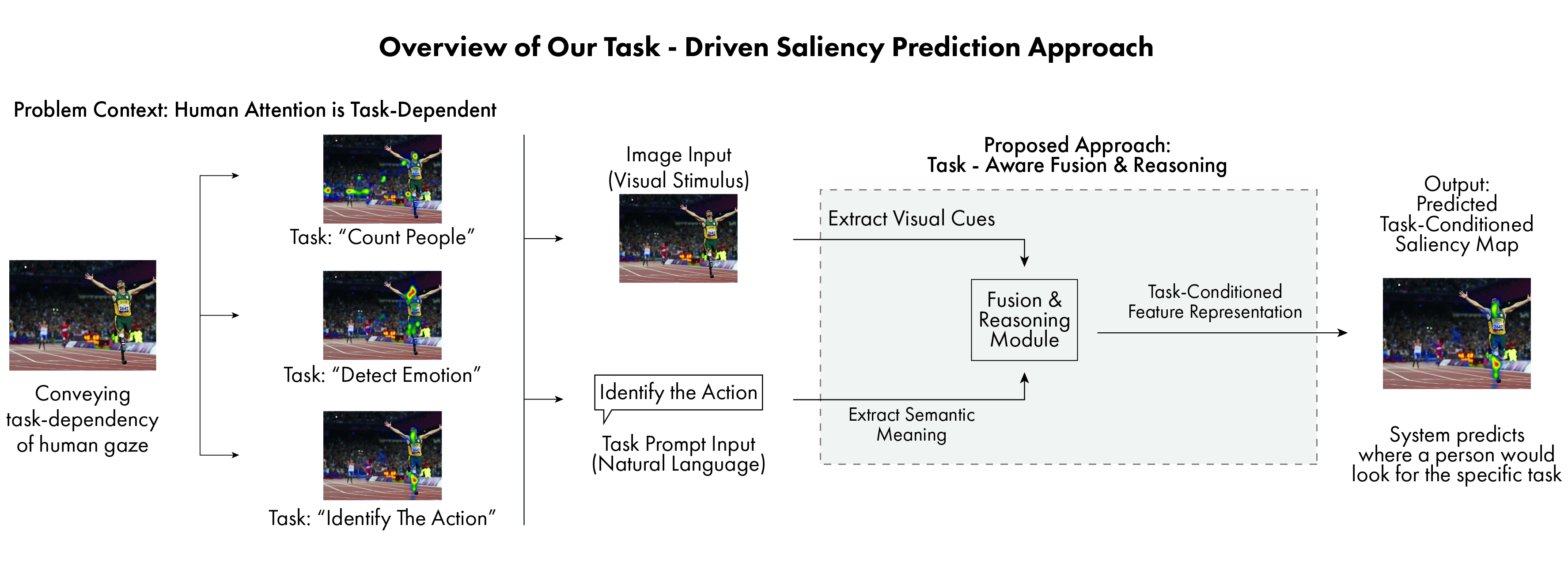}
    \caption{Conceptual overview of task-driven saliency prediction. 
    Different task prompts lead to distinct human fixation patterns on the same image. We model this by fusing visual features extracted from the image with semantic representations of the task prompt, producing a task-based saliency map that reflects where a viewer would look given a specific goal.}
    \label{fig:overview}
\end{figure}

Figure~\ref{fig:overview} provides an overview of our task-driven formulation. The model combines (i) spatial visual features extracted from the input image and (ii) a semantic representation of the task prompt, and outputs a task-conditioned saliency map. This schematic summarizes the core fusion pathway that is detailed in Section~\ref{sec:method}.

\subsection{Contributions}

Our work introduces \textbf{TDSal}, a modular vision-language architecture for task-driven saliency prediction. Building upon the research questions outlined above, our main contributions are as follows:

\begin{itemize}
  \item We introduce \textbf{TDSal}, a modular top-down saliency prediction network that fuses YOLO-derived spatial visual features with Sentence-BERT task embeddings via a transformer-based fusion block.
  \item We propose an architectural design for task-driven saliency, where semantic conditioning is integrated through a shallow fusion transformer used exclusively for cross-modal interaction, rather than through deep multimodal backbones.
  \item We validate our model through seven saliency metrics (CC, KLDiv, SIM,  NSS, AUC--Borji, AUC-J, and sAUC)~\cite{kummerer2018saliency,borji2012quantitative}  and qualitative visualizations, and position TDSal against related task-driven  and top-down saliency models (TGSal~\cite{sun2024visual} and  SalClassNet~\cite{murabito2018top}), the closest available dense saliency  references with task-driven or top-down conditioning.
  \item We provide an analysis showing that architectural choices affect different saliency properties in different ways, highlighting the need to evaluate task-driven saliency beyond a single aggregate score.
\end{itemize}

\section{Related Work}
\label{sec:related}

Visual saliency aims to predict where humans fixate within a scene. Traditional bottom-up models rely on low-level cues such as color or contrast and assume task-free viewing, whereas top-down saliency reflects how goals and semantic intent shape attention. Task-driven saliency research therefore focuses on how high-level intent influences gaze behavior, often using eye-tracking datasets whose fixation density maps (FDMs) provide probabilistic ground truth under different task conditions. This perspective motivates models that can integrate semantic cues with visual structure, forming the basis for the approaches reviewed below.

Research in saliency prediction has evolved from purely bottom-up methods to top-down approaches that incorporate task-specific and semantic cues. In this section, we categorize prior work into three main directions: (1) early bottom-up and task-free saliency models, (2) task-driven saliency prediction, and (3) vision–language fusion methods for attention modeling.

\subsection{Bottom-Up vs. Task-Driven Saliency Models}
Early saliency research primarily focused on bottom-up models that rely on low-level features such as color, texture, and contrast. These models assume a universal attention pattern, independent of viewer intent. Learning-based saliency models later showed that combining low, mid, and high-level cues learned from eye-tracking data improves fixation prediction over purely hand-crafted bottom-up cues~\cite{judd2009learning}. However, such approaches still fail to capture task-based behaviors present in many real-world scenarios.

An early attempt at top-down saliency was made by Liang and Zhang~\cite{liang2015topdown}, who introduced a KL-divergence–based formulation to identify regions contributing most to object recognition. Although their approach conditions attention on semantics indirectly through class distributions, it does not incorporate explicit task or goal definitions.

Top-down models address this limitation by conditioning attention on high-level intent or task definitions. For example, Murabito et al.~\cite{murabito2018top} propose saliency maps that emerge from classification tasks, aligning attention with class-relevant regions. Similarly, Albayrak~\cite{albayrak2020study} explores task-conditioned attention using eye-tracking under various viewing goals. These models demonstrate that task semantics can significantly alter attention patterns.

Motivated by these findings, we design TDSal to incorporate task semantics explicitly while retaining a spatially rich visual backbone for saliency prediction.

\subsection{Saliency Prediction Using Transformers}
Transformers have shown strong performance in saliency prediction by modeling global dependencies. Liu et al.'s Visual Saliency Transformer (VST)~\cite{liu2021visual} uses patch-wise tokenization and multihead self-attention to capture long-range interactions, improving over convolutional baselines. While effective, these methods often require large-scale data and computation, making them less suitable for real-time applications.

Transformers have also been used in multimodal settings, where visual tokens are combined with textual or semantic tokens to guide spatial attention. This formulation enables cross-modal interactions, allowing linguistic cues to influence the weighting and interpretation of spatial features. Prior work in vision-language models has shown that appending or concatenating textual embeddings as additional tokens within a transformer sequence can effectively steer attention toward regions aligned with linguistic intent~\cite{wen2024efficient,neo2024towards}.

Our work draws inspiration from this direction but simplifies the design by using a transformer encoder only for fusion, while relying on YOLO for efficient spatial feature extraction. In contrast to large transformer-based saliency models that require extensive computation, our approach limits transformer usage to multimodal fusion, maintaining the contextual reasoning needed for task-driven attention without architectural complexity.

\subsection{Vision–Language Fusion for Attention Modeling}

Recent models have increasingly bridged vision and language to enable conditioned attention. Sentence-BERT~\cite{reimers2019sentence} and other pretrained language encoders convert textual prompts into dense semantic embeddings suitable for multimodal learning. Early work by Ramanishka et al.~\cite{Ramanishka_2017_CVPR} demonstrated that natural language captions can effectively guide visual saliency, laying the foundation for vision–language grounded attention modeling. 

Building upon this work, MDETR~\cite{kamath2021mdetr} employs a transformer-based architecture to align textual queries with object-level visual regions for modulated detection and grounding. While it demonstrates effective vision--language alignment, it operates at the bounding-box level and does not target dense fixation or saliency prediction. More recently, TDiffSal~\cite{zhang2024tdiffsal} adopts a diffusion-based framework to generate saliency maps directly conditioned on textual descriptions, highlighting the growing relevance of text-driven saliency prediction.

In contrast to models primarily designed for detection (e.g., MDETR~\cite{kamath2021mdetr}) or computationally heavy generative frameworks (e.g., TDiffSal~\cite{zhang2024tdiffsal}), we focus on \emph{dense} task-driven saliency prediction with a fusion design. TDSal combines a compact Sentence-BERT task embedding with spatial visual features from an object-detection backbone, and uses a shallow transformer to translate task intent into structured spatial emphasis in the predicted saliency map.

TDSal builds on these directions by combining language-based task conditioning with object-centered visual features for dense saliency prediction.

\begin{figure}[t]
    \centering
    \includegraphics[width=\textwidth]{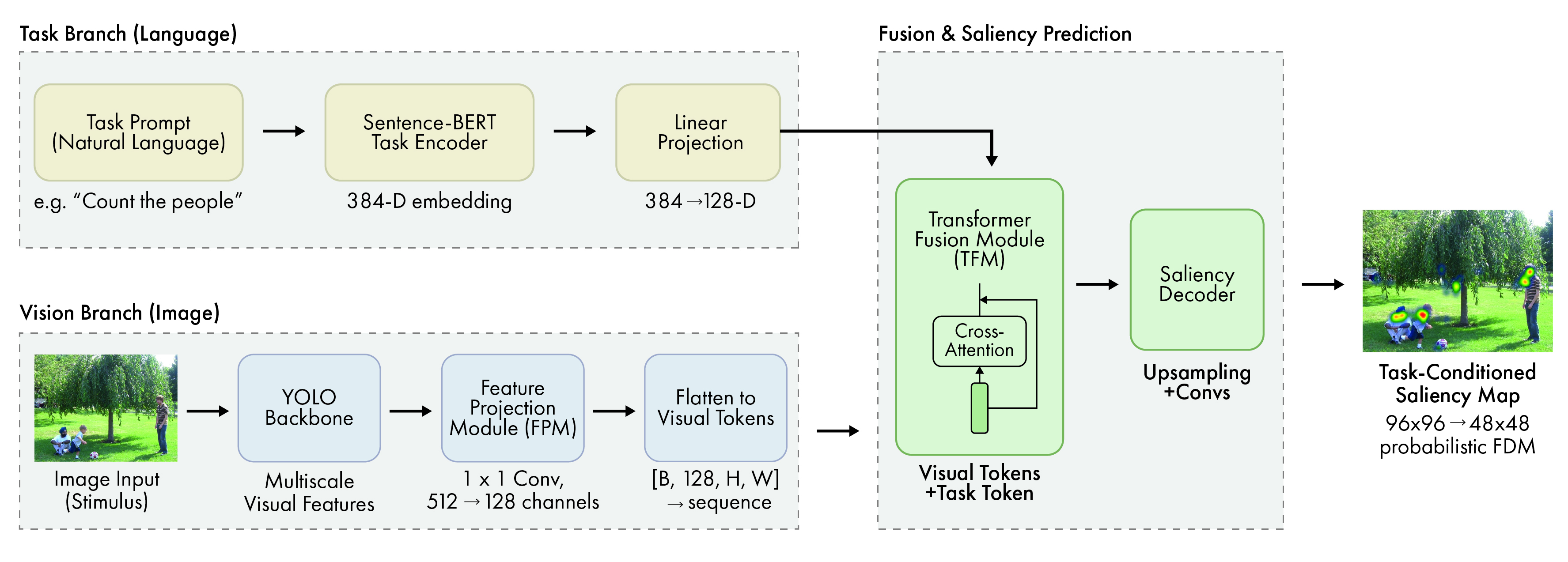}
    \caption{Overview of the model architecture.}
    \label{fig:model_architecture}
\end{figure}

\section{Proposed Method}
\label{sec:method}
We propose \textbf{TDSal}, a task-conditioned saliency model composed of four main stages: 
(i) a YOLO-based backbone that extracts spatial visual feature maps, 
(ii) a $1{\times}1$ Feature Projection Module (FPM) that reduces channel dimensionality, 
(iii) a Sentence-BERT task encoder that produces a compact task embedding, and 
(iv) a shallow Transformer Fusion Module (TFM) that integrates the task token with spatial visual tokens. 
A decoder then reconstructs the final saliency map.

\subsection{Design Hypotheses}

The architectural design of TDSal is guided by the following hypotheses, which aim to answer the research questions stated in Section~\ref{sec:intro}:

\begin{itemize}
    \item Incorporating an object detection backbone such as YOLO enables the model to extract semantically rich and spatially precise visual features that align with human gaze patterns under specific tasks.
    \item Encoding high-level task definitions through Sentence-BERT embeddings provides meaningful semantic cues that can steer saliency prediction toward goal-relevant regions.
    \item Fusing visual and textual representations provides an explicit mechanism for task-aware attention modulation, enabling the model to condition saliency predictions on high-level viewing intent.
\end{itemize}

\subsection{Model Architecture}

The architecture consists of a YOLO-based backbone for visual feature extraction, an FPM (1×1 conv projection), a task encoder to generate semantic embeddings, a transformer fusion module that combines visual and semantic features, and a final saliency decoder that produces the output saliency maps (see Figure~\ref{fig:model_architecture}).

\subsubsection{YOLO Backbone}
The backbone employs the first 10 layers (layers~0--9) of a pre-trained YOLOv5su model~\cite{redmon2016you,yolov5}, truncated at the SPPF (Spatial Pyramid Pooling -- Fast) module and discarding the detection head. For a $384{\times}384$ input, this yields a \emph{single} feature map of shape $[B,\,512,\,12,\,12]$ at stride-32. No multi-scale feature pyramid is constructed prior to the FPM; the SPPF module itself provides implicit multi-receptive-field aggregation by applying max-pooling at kernel sizes $5{\times}5$, $9{\times}9$, and $13{\times}13$ and concatenating the results, which is the sense in which we refer to spatially rich visual features. This single-scale output is passed directly to the FPM. YOLO's hierarchical architecture captures both low-level cues such as edges and textures and higher-level semantic structures such as objects and contextual relationships, which are known to align well with human visual attention~\cite{redmon2016you,yolov5}. Leveraging the early and intermediate layers provides an object-centered prior that enhances scene understanding while remaining efficient, since the features are optimized for representation learning rather than final classification.

\subsubsection{Feature Projection Module (FPM)}

The Feature Projection Module applies a single $1{\times}1$ convolution to the high-dimensional feature maps produced by the YOLO backbone, reducing their channel dimensionality from 512 to 128 while preserving spatial resolution. This operation serves to distill the most informative visual cues into a more compact representation that retains critical spatial details. The reduced dimensionality minimizes computational cost and memory footprint during transformer fusion, ensuring that only the most salient and task-relevant features are carried forward. By simplifying the feature space without losing semantic richness, the FPM enables efficient cross-modal fusion and contributes to the overall effectiveness of the saliency decoder.

\subsubsection{Task Encoder}
To incorporate task-specific context, the task encoder converts textual task descriptions into dense semantic embeddings using Sentence-BERT~\cite{reimers2019sentence}. We use the MiniLM-L6 SentenceTransformer variant, which produces a 384-dimensional sentence embedding for each task prompt. This embedding is passed through a linear projection layer to produce a 128-dimensional task token, matching the dimensionality of the visual tokens produced by the Feature Projection Module. This alignment allows the task token to be appended directly to the visual-token sequence before transformer fusion. The resulting representation bridges the language--vision gap, enabling semantic task intent to guide spatial attention toward image regions that are most relevant to the described goal.

\subsubsection{Transformer Fusion Module (TFM)}
The Transformer Fusion Module unifies the visual and semantic representations to enable task-aware attention. Feature maps from the FPM are first flattened into a sequence of visual tokens, while the encoded task embedding expressing the high-level intent of the task is appended as an additional token. This combined sequence is processed by a transformer encoder that leverages self-attention to model the interactions between task semantics and spatial image features. Through this mechanism, the module learns to highlight regions that are most relevant to the specified task, effectively translating semantic guidance into spatial focus. The fused representation is then reshaped back into spatial form and forwarded to the saliency decoder for reconstruction.

\subsubsection{Saliency Decoder}

The saliency decoder reconstructs a dense attention map from the fused multimodal representation produced by the transformer. Starting from the fused $12 \times 12$ feature map, the decoder applies a shallow sequence of convolutional and upsampling layers that progressively restores spatial resolution to $96 \times 96$. This design preserves spatial structure while avoiding unnecessary architectural depth or computational cost. A final sigmoid activation bounds the output values to the $[0,1]$ range. During training, the predicted $96 \times 96$ map is bilinearly downsampled to $48 \times 48$ to match the ground-truth fixation density maps before computing the loss. For distribution-based losses and metrics such as KLDiv and SIM, both predicted and ground-truth maps are normalized over spatial locations before computation.

\begin{table}[htbp]
  \centering
  \caption{Implementation Details}
  \label{tab:impl_details}
  \footnotesize
  \setlength{\tabcolsep}{3pt}
  \renewcommand{\arraystretch}{1.15}

  \begin{tabularx}{\columnwidth}{|p{2.1cm}|Y|}
    \hline
    \textbf{Component} & \textbf{Key Settings} \\ \hline
    Backbone & YOLOv5su (Ultralytics)\\
             & Single output: $[B,512,12,12]$, stride-32\\
             & first 10 layers (SPPF 512-ch) \\ \hline
    FPM Module & $1{\times}1$ conv: 512 $\to$ 128 ch \\
               & output $[B,128,H,W]$ \\ \hline
    Task Encoder & MiniLM-L6 SentenceTransformer: 384 $\rightarrow$ 128 task token \\ \hline
    Transf. Fusion & 1-layer Encoder ($d$=128, $h$=4) + 1 task token \\ \hline
    Saliency Dec. & Conv/upsampling decoder: $12{\times}12 \rightarrow 96{\times}96$ \\
    & Sigmoid output: $[B,1,96,96]$ \\ \hline
    Reproducibility & Random seed 42 \\ \hline
  \end{tabularx}
\end{table}

\subsection{Dataset \& Preprocessing}
The dataset used here was originally collected by Albayrak~\cite{albayrak2020study}. The dataset uses image stimuli sourced from the Emotional Attention dataset~\cite{fan2018emotional} and the Saliency in Crowd dataset~\cite{jiang2014saliency}.

Large-scale saliency benchmarks such as SALICON have enabled data-driven saliency modeling by providing scalable human attention annotations (via mouse-tracking proxies)~\cite{jiang2015salicon}. In contrast, this dataset provides eye-tracking fixation density maps collected under four explicit task conditions, where each task is defined by a natural-language viewing goal.

In our experiments, we use the raw grayscale maps as ground truth, yielding a total of 1,968 image–saliency pairs across the four tasks. We split these into:
\begin{itemize}
  \item 70\% train (1,377 pairs)  
  \item 15\% validation (295 pairs)  
  \item 15\% test (296 pairs)  
\end{itemize}
All splits are generated with a fixed random seed (42) to ensure reproducibility.

Preprocessing and augmentations are applied identically to each image–map pair:
\begin{itemize}
  \item \textbf{Resize:} Stimuli to 384\,$\times$\,384 px. Ground-truth high-level FDMs are recorded at 48\,$\times$\,48; we keep them at that size and later down-sample predictions to 48\,$\times$\,48 before computing the loss. 
  \item \textbf{Normalization:} scale image pixel values to $[0,1]$ and convert to tensors.  
  \item \textbf{Paired augmentations:} random horizontal flip (probability 0.5) and random rotation within $\pm$10°, applied the same way to the image and its corresponding map to preserve spatial correspondence.
\end{itemize}

Ground-truth fixation density maps are recorded at a reduced spatial resolution. We keep them at this resolution and down-sample predictions accordingly before computing the loss, consistent with standard saliency evaluation, where fixation maps are defined as smoothed density maps~\cite{kummerer2018saliency}.

\subsection{Training Configuration}
We train our TDSal model for 50 epochs. The optimizer is Adam with a constant learning rate of $1\times10^{-4}$ and no additional weight decay. We use a batch size of 8 and reset gradients at each step to stabilize multimodal fusion on a modest dataset without overfitting.

The loss is a combined saliency objective:
\[
\mathcal{L}(S,\hat{S}) =
\alpha \, KL(S \parallel \hat{S})
+
\beta \, \bigl(1 - CC(S,\hat{S})\bigr),
\]
with weighting factors $\alpha=\beta=1.0$. Before computing distribution-based terms, both the predicted saliency map $\hat{S}$ and the ground-truth fixation density map $S$ are normalized over spatial locations. KLDiv measures the discrepancy between the predicted and ground-truth saliency distributions, while CC measures their spatial correlation~\cite{kummerer2018saliency}.

At each forward pass, the model outputs a 96\,$\times$\,96 saliency map, which we bilinearly downsample to 48\,$\times$\,48 to match the ground-truth eye-tracker maps before computing the loss. We log both batch-level and epoch-average loss values to monitor convergence (Figure~\ref{fig:training-validation-trends}(a)).

\section{Results}
\label{sec:results}
\subsection{Quantitative Evaluation}
To evaluate the performance of TDSal, we report five widely adopted saliency metrics across the validation and test splits: Pearson's Correlation Coefficient (CC) ~\cite{kummerer2018saliency}, Kullback--Leibler Divergence (KLDiv)~\cite{liang2015topdown,kummerer2018saliency}, Similarity (SIM)~\cite{kummerer2018saliency}, Normalized Scanpath Saliency (NSS)~\cite{kummerer2018saliency}, and AUC--Borji~\cite{borji2012quantitative}. These metrics reflect different aspects of saliency quality: CC measures spatial correlation, KLDiv measures distribution divergence, SIM captures global map similarity, NSS evaluates alignment with fixation locations, and AUC--Borji measures the discriminative power between salient and non-salient regions under a random-sampling protocol.

In addition, AUC-J and shuffled AUC (sAUC) are reported in the literature-context and ablation analyses to provide broader coverage of ranking-based saliency behavior.

\begin{table}[htbp]
  \caption{Validation vs. Test performance across five saliency metrics.}
  \label{tab:final-metrics}
  \centering
  \footnotesize
  \renewcommand{\arraystretch}{1.15}
  \begin{tabular}{|l|c|c|}
    \hline
    \textbf{Metric} & \textbf{Validation} & \textbf{Test} \\ \hline
    CC         & 0.6516 & 0.6423 \\ 
    KLDiv      & 0.9132 & 0.9270 \\ 
    SIM        & 0.5006 & 0.5010 \\ 
    NSS        & 3.6885 & 3.4583 \\ 
    AUC--Borji  & 0.9549 & 0.9177 \\ \hline
  \end{tabular}
\end{table}

As shown in Table~\ref{tab:final-metrics}, validation and test metrics are closely aligned across all measures, with differences typically below 6\%, indicating strong generalization and no signs of overfitting.

Our model achieves an NSS of 3.46 and an AUC--Borji of 0.9177 under the task-conditioned evaluation protocol. These values indicate strong fixation alignment within the proposed experimental setting. Since the dataset, task formulation, and annotation protocol differ from free-viewing benchmarks such as MIT300 and SALICON, we do not interpret these scores as direct benchmark comparisons.

We follow the saliency evaluation methodology outlined by Kümmerer et al.~\cite{kummerer2018saliency}, which emphasizes separating models, maps, and metric computation. Accordingly, the reported values are used to evaluate TDSal under a consistent task-conditioned protocol rather than to claim direct superiority over models evaluated on different datasets.

\begin{figure}[t]
  \centering
  \begin{minipage}{0.48\textwidth}
    \centering
    \includegraphics[width=\linewidth]{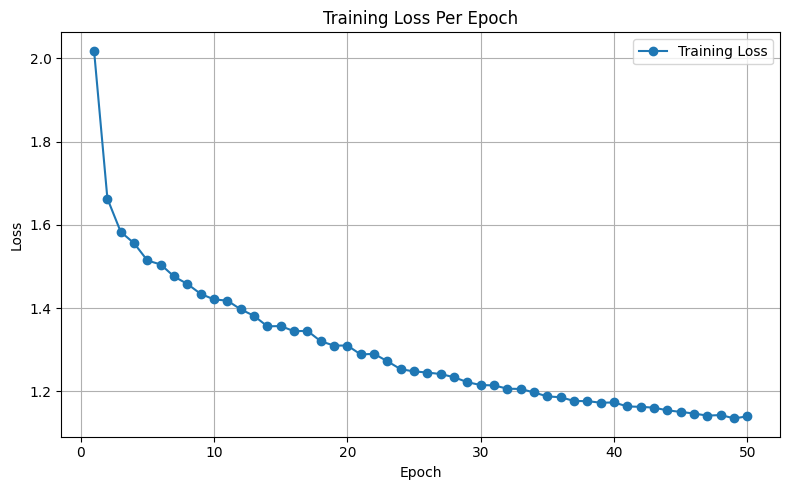}
    \vspace{-1mm}
    \small (a) Training loss
  \end{minipage}
  \hfill
  \begin{minipage}{0.51\textwidth}
    \centering
    \includegraphics[width=\linewidth]{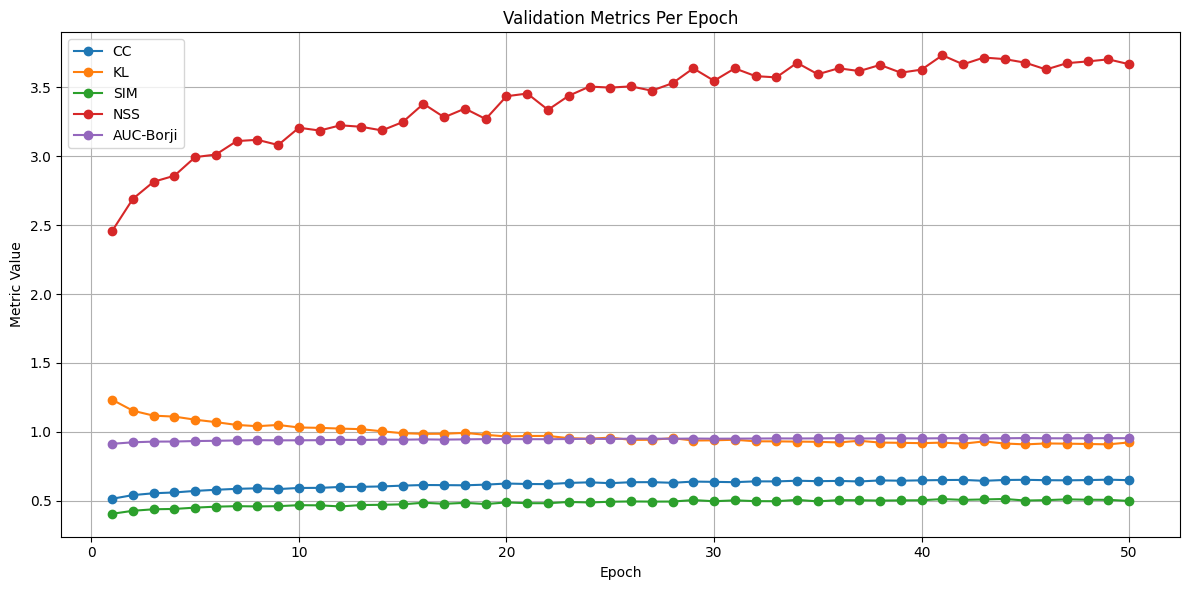}
    \vspace{-1mm}
    \small (b) Validation metrics
  \end{minipage}
  \caption{Training and validation trends over 50 epochs.}
  \label{fig:training-validation-trends}
\end{figure}

\subsection{Training and Validation Trends}

Figure~\ref{fig:training-validation-trends} shows stable optimization over 50 epochs. The training loss decreases from above 2.00 to approximately 1.18 without major oscillations. At the same time, validation NSS and AUC--Borji increase steadily, while CC and SIM improve gradually and KL divergence decreases. These trends indicate stable convergence and progressively improved alignment with task-conditioned fixation distributions.

\subsection{Qualitative Visualization}

\begin{figure}[t]
    \centering
    \includegraphics[width=\textwidth]{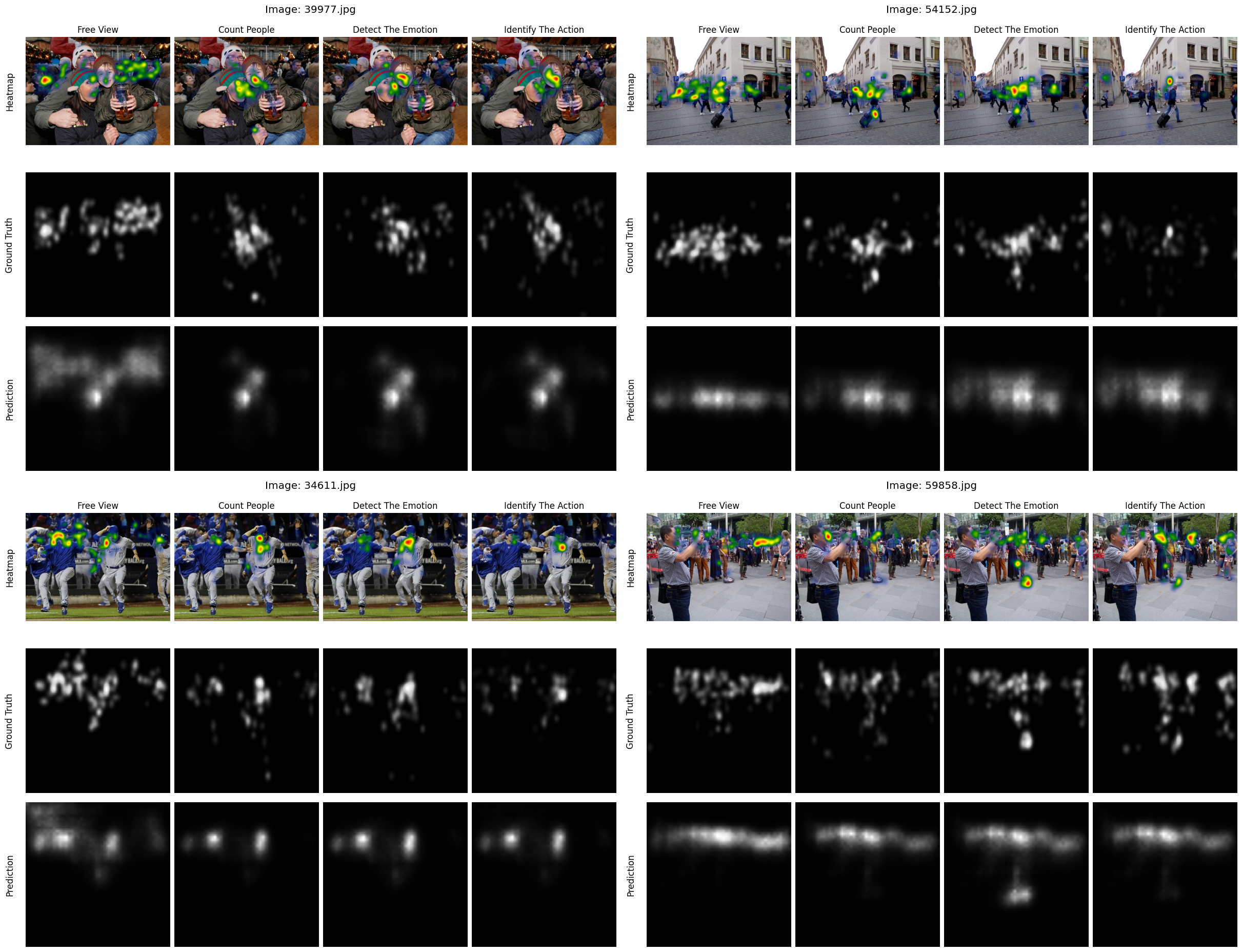}
    \caption{Task-conditioned saliency predictions. Rows show input overlays, ground-truth fixation density maps, and predicted saliency maps.}
    \label{fig:qualitative-results}
\end{figure}

Figure~\ref{fig:qualitative-results} shows predictions for four images under different task prompts. The predicted maps shift according to the prompt, indicating that TDSal uses task semantics to modulate spatial attention while remaining grounded in the image content.

\subsection{Comparison with Task-Driven and Top-Down Saliency Models in Literature}
\label{sec:task_comparison}
Since TDSal is evaluated on a task-conditioned eye-tracking dataset, free-viewing saliency benchmarks such as MIT300 and SALICON are not treated as direct experimental baselines because they differ in stimuli, annotation protocol, task formulation, and dataset bias. We therefore focus on related task-driven or top-down saliency-map models that report overlapping metrics, while interpreting all values as contextual references rather than strict performance and quality comparisons.

We include TGSal~\cite{sun2024visual} and SalClassNet~\cite{murabito2018top} because both produce dense saliency maps and report partially overlapping saliency metrics. TGSal uses image captions or descriptions rather than explicit task prompts, while SalClassNet is driven by visual classification rather than natural-language task descriptions. Gazeformer~\cite{mondal2023gazeformer} and AiR~\cite{chen2020air} are excluded from the table because they evaluate scanpaths or reasoning-step attention rather than dense saliency maps.

\begin{table}[t]
  \centering
  \caption{Reported values from related task-driven and top-down saliency models. Values are contextual references, not direct benchmark comparisons.}
  \label{tab:task-driven-comparison}
  \scriptsize
  \renewcommand{\arraystretch}{1.0}
  \setlength{\tabcolsep}{2.2pt}

  \begin{tabular}{|l|c|c|c|c|c|c|c|}
    \hline
    \textbf{Model} 
    & \textbf{\makecell{AUC--\\Borji}} 
    & \textbf{AUC-J}
    & \textbf{sAUC}
    & \textbf{CC} 
    & \textbf{KLDiv} 
    & \textbf{SIM} 
    & \textbf{NSS} \\
    \hline
    TDSal & 0.9177 & 0.9515 & 0.8649 & 0.6423 & 0.9270 & 0.5010 & 3.4583 \\
    TGSal~\cite{sun2024visual}
    & -- & 0.8717 & 0.6691 & 0.7700 & 6.3760\textsuperscript{a} & 0.6847 & 1.9409 \\
    SalClassNet~\cite{murabito2018top}
    & -- & -- & 0.862 & 0.461 & -- & -- & 4.239 \\
    \hline
  \end{tabular}

  \vspace{0.5mm}
  \begin{minipage}{0.95\textwidth}
  \scriptsize
  \textsuperscript{a} KLDiv values are not directly comparable across protocols; TGSal follows the SJTU-TIS evaluation scale. Missing values were not reported in the original papers.
  \end{minipage}
\end{table}

Table~\ref{tab:task-driven-comparison} shows that TDSal obtains strong fixation-alignment and ranking-based values under the task-conditioned evaluation setting, particularly for NSS, AUC-J, sAUC, and AUC--Borji. However, these values should not be interpreted as a strict comparison: TDSal uses explicit natural-language task prompts, TGSal uses image captions or descriptions, and SalClassNet uses a visual classification objective. Their datasets and evaluation protocols also differ. The table therefore highlights methodological proximity and metric coverage, while motivating the need for standardized benchmarks for natural-language task-conditioned dense saliency prediction.

\subsection{Ablation Study}

The ablation study is conducted to analyze how individual architectural components contribute to different properties of task-driven visual attention, rather than to optimize a single evaluation metric. 
Following prior saliency evaluation protocols, we group the metrics into three functional categories:
(i) \textit{spatial correlation} with human attention (CC),
(ii) \textit{distributional alignment} (KL divergence and AUC--Borji), and
(iii) \textit{saliency strength} (NSS and SIM).

\begin{table}[t]
  \centering
  \caption{Ablation study on TDSal.}
  \label{tab:ablation}
  \scriptsize
  \setlength{\tabcolsep}{2.2pt}
  \renewcommand{\arraystretch}{1.0}
  \begin{tabular}{|l|c|c|c|c|c|c|c|}
    \hline
    \textbf{Model}
    & \textbf{CC}
    & \textbf{KL}
    & \textbf{SIM}
    & \textbf{NSS}
    & \textbf{\makecell{AUC--\\Borji}}
    & \textbf{AUC-J}
    & \textbf{sAUC} \\
    \hline
    Full TDSal      & 0.6435 & 0.9600 & 0.5160 & 3.6261 & 0.9011 & 0.9477 & 0.8630 \\
    w/o Task        & 0.6508 & 0.9661 & 0.5263 & 3.6776 & 0.8984 & 0.9496 & 0.8637 \\
    w/o Transformer & 0.6902 & 0.8675 & 0.5575 & 3.9983 & 0.9188 & 0.9599 & 0.8854 \\
    w/o SBERT       & 0.6375 & 1.0347 & 0.5267 & 3.7646 & 0.8682 & 0.9495 & 0.8629 \\
    w/o FPM         & 0.6481 & 0.9504 & 0.5263 & 3.7330 & 0.8904 & 0.9508 & 0.8641 \\
    \hline
  \end{tabular}
\end{table}

The ablation results show that different architectural components affect different saliency properties. Some ablated variants obtain higher aggregate scores for metrics such as NSS, SIM, or CC. This can occur when predictions become sharper or more globally correlated with fixation density maps, but such gains do not necessarily imply better task-driven saliency. In particular, aggregate saliency metrics may not fully capture whether attention shifts are semantically consistent with the given task prompt.

The full TDSal model preserves the complete task-conditioned formulation, including explicit language input, projected task-token representation, transformer-based cross-modal fusion, and feature projection before decoding. Although the full model does not dominate every metric in the ablation table, it provides the most semantically complete architecture for modeling task-driven visual attention. These results suggest that future evaluation should include task-specific and semantic-consistency analyses in addition to aggregate saliency metrics.

Therefore, the ablation study should be interpreted as an indicative component analysis rather than as proof that every module independently improves every metric. Each variant is evaluated from a single trained checkpoint, and the observed differences may reflect metric-specific trade-offs between sharpness, distributional alignment, and task-conditioned semantic coherence.

\section{Discussion}

\subsection{Limitations}

Despite the strong quantitative and qualitative results, several limitations remain. First, the dataset used in this study comprises 1,968 image–task pairs across four fixed task categories. While sufficient to demonstrate the feasibility and effectiveness of task-driven saliency modeling, this limited scale and semantic diversity may constrain the model’s ability to generalize to unseen task formulations, open-ended natural language prompts, or domain-shifted visual content.

Second, although we report an ablation study that analyzes the contribution of  key architectural components, the analysis remains metric-driven and  task-aggregated. We do not explicitly examine task-specific ablation effects  (e.g., how different components influence performance under individual task  types), nor do we analyze cross-task transfer behavior. Additionally, each  ablation variant corresponds to a single trained checkpoint, so the observed  metric differences should be treated as indicative trends rather than  statistically confirmed results. As a result, finer-grained insights into how  architectural elements support specific task semantics remain unexplored.

Third, cross-dataset generalization could not be evaluated on benchmarks such as SALICON or MIT300 due to the lack of task-conditioned eye-tracking annotations on those datasets. Consequently, while our results demonstrate strong alignment with human attention under controlled task settings, they do not directly establish robustness across heterogeneous saliency benchmarks or free-viewing scenarios. As saliency evaluation is known to be sensitive to dataset bias and metric choice, absolute performance comparisons across datasets should therefore be interpreted with caution~\cite{borji2012quantitative}.

\subsection{Future Work}

Future work will focus on extending both the empirical scope and analytical depth of TDSal. A primary direction is the expansion of the existing dataset through additional eye-tracking data collection, with the goal of increasing task diversity and improving generalization to more complex or compositional task descriptions.

From a modeling perspective, future work will extend the current ablation analysis by conducting task-specific and cross-task studies, enabling a deeper understanding of how individual components contribute under different semantic conditions. This includes examining whether components such as the transformer fusion module or the task encoder provide consistent benefits across all task types or primarily support certain forms of task-driven attention.

Beyond metric-level evaluation, future work will assess model robustness under challenging visual conditions, such as cluttered scenes, low-contrast stimuli, and visually ambiguous task scenarios. The current qualitative evaluation demonstrates task-driven attention shifts under controlled conditions, but does not systematically probe failure modes or edge cases that may arise in more complex real-world settings.

Architecturally, future work will explore multi-layer transformer fusion, task-adaptive decoding strategies, and alternative language encoders to better capture task semantics. Finally, we aim to investigate cross-dataset evaluation strategies, such as weakly supervised transfer or synthetic task conditioning, to assess robustness beyond the current dataset without requiring full task-specific eye-tracking annotations.

A further direction is the establishment of direct same-dataset comparisons with related task-driven models such as TGSal and SalClassNet. Currently, the lack of a shared benchmark and publicly available checkpoints for these models prevents evaluation on the same data split. Future work will explore whether such comparisons become feasible as standardized task-conditioned saliency benchmarks emerge in the community.

\section{Conclusion}

We presented TDSal, a top-down saliency prediction network that combines the first 10 layers of a pre-trained YOLOv5su backbone, a $1{\times}1$ Feature Projection Module, a Sentence-BERT task encoder, and a transformer-based fusion block. On a task-oriented eye-tracking dataset, TDSal attains an NSS of 3.46, AUC--Borji of 0.9177, AUC-J of 0.9515, and sAUC of 0.8649 on the test split. Because the evaluation is task-conditioned, these values are not directly comparable to MIT/Tübingen-style free-viewing benchmarks; they should instead be interpreted as evidence of fixation alignment within the controlled task-driven setting used in this work.

Quantitative trends and qualitative visualizations indicate that TDSal shifts attention in line with task prompts, supporting the value of combining high-level semantic intent with visual features for goal-directed saliency prediction. At the same time, the ablation results show that aggregate saliency metrics do not always favor the full model uniformly, suggesting that future work should include task-specific and semantic-consistency evaluations in addition to standard saliency metrics.

While the current evaluation is limited by dataset scale and the absence of standardized task-conditioned benchmarks, the results show that explicit task conditioning with multimodal fusion is a promising direction for modeling goal-directed visual attention.

\begin{credits}

\subsubsection{\discintname}
The authors have no competing interests to declare that are relevant to the content of this article.
\end{credits}

\bibliographystyle{splncs04}
\bibliography{references}

@mastersthesis{albayrak2020study,
    author = {Albayrak, Dilara},
    title = {A Study of Visual Saliency for Free-Viewing and Task-Oriented Condition},
    school = {TED University},
    year = {2020},
    url    = {https://github.com/DilaraAlbayrak/Task-based-eye-fixation-dataset}
}

@inproceedings{liu2021visual,
  title={Visual saliency transformer},
  author={Liu, Nian and Zhang, Ni and Wan, Kaiyuan and Shao, Ling and Han, Junwei},
  booktitle={Proceedings of the IEEE/CVF international conference on computer vision},
  pages={4722--4732},
  year={2021}
}

@article{murabito2018top,
  title={Top-down saliency detection driven by visual classification},
  author={Murabito, Francesca and Spampinato, Concetto and Palazzo, Simone and Giordano, Daniela and Pogorelov, Konstantin and Riegler, Michael},
  journal={Computer Vision and Image Understanding},
  volume={172},
  pages={67--76},
  year={2018},
  publisher={Elsevier}
}

@inproceedings{kummerer2018saliency,
  title={Saliency benchmarking made easy: Separating models, maps and metrics},
  author={Kummerer, Matthias and Wallis, Thomas SA and Bethge, Matthias},
  booktitle={Proceedings of the European Conference on Computer Vision (ECCV)},
  pages={770--787},
  year={2018}
}

@INPROCEEDINGS{liang2015topdown,
  author={Liang, Jun and Zhang, Yanning},
  booktitle={2015 International Symposium on Bioelectronics and Bioinformatics (ISBB)}, 
  title={Top down saliency detection via Kullback-Leibler divergence for object recognition}, 
  year={2015},
  volume={},
  number={},
  pages={200-203},
  doi={10.1109/ISBB.2015.7344958}}

@article{borji2012quantitative,
  title={Quantitative analysis of human-model agreement in visual saliency modeling: A comparative study},
  author={Borji, Ali and Sihite, Dicky N and Itti, Laurent},
  journal={IEEE Transactions on Image Processing},
  volume={22},
  number={1},
  pages={55--69},
  year={2012},
  publisher={IEEE}
}

@inproceedings{redmon2016you,
  title={You only look once: Unified, real-time object detection},
  author={Redmon, Joseph and Divvala, Santosh and Girshick, Ross and Farhadi, Ali},
  booktitle={Proceedings of the IEEE conference on computer vision and pattern recognition},
  pages={779--788},
  year={2016}
}

@InProceedings{Ramanishka_2017_CVPR,
author = {Ramanishka, Vasili and Das, Abir and Zhang, Jianming and Saenko, Kate},
title = {Top-Down Visual Saliency Guided by Captions},
booktitle = {Proceedings of the IEEE Conference on Computer Vision and Pattern Recognition (CVPR)},
month = {July},
year = {2017}
}

@inproceedings{zhang2024tdiffsal,
  title={TDiffSal: Text-Guided Diffusion Saliency Prediction Model for Images},
  author={Zhang, Nana and Xiong, Min and Zhu, Dandan and Zhu, Kun and Zhai, Guangtao},
  booktitle={International Conference on Pattern Recognition},
  pages={15--31},
  year={2024},
  organization={Springer}
}

@inproceedings{kamath2021mdetr,
  title={Mdetr-modulated detection for end-to-end multi-modal understanding},
  author={Kamath, Aishwarya and Singh, Mannat and LeCun, Yann and Synnaeve, Gabriel and Misra, Ishan and Carion, Nicolas},
  booktitle={Proceedings of the IEEE/CVF international conference on computer vision},
  pages={1780--1790},
  year={2021}
}

@article{reimers2019sentence,
  title={Sentence-bert: Sentence embeddings using siamese bert-networks},
  author={Reimers, Nils and Gurevych, Iryna},
  journal={arXiv preprint arXiv:1908.10084},
  year={2019}
}

@misc{yolov5,
  author       = {Glenn Jocher and Ultralytics},
  title        = {Ultralytics {YOLOv5}},
  year         = {2020},
  howpublished = {\url{https://github.com/ultralytics/yolov5}},
  note         = {Version 7.0},
  doi          = {10.5281/zenodo.3908559}
}

@article{neo2024towards,
  title={Towards interpreting visual information processing in vision-language models},
  author={Neo, Clement and Ong, Luke and Torr, Philip and Geva, Mor and Krueger, David and Barez, Fazl},
  journal={arXiv preprint arXiv:2410.07149},
  year={2024}
}

@article{wen2024efficient,
  title={Efficient vision-language models by summarizing visual tokens into compact registers},
  author={Wen, Yuxin and Cao, Qingqing and Fu, Qichen and Mehta, Sachin and Najibi, Mahyar},
  journal={arXiv preprint arXiv:2410.14072},
  year={2024}
}

@inproceedings{judd2009learning,
  title={Learning to predict where humans look},
  author={Judd, Tilke and Ehinger, Krista and Durand, Fr{\'e}do and Torralba, Antonio},
  booktitle={2009 IEEE 12th international conference on computer vision},
  pages={2106--2113},
  year={2009},
  organization={IEEE}
}

@inproceedings{jiang2015salicon,
  title={Salicon: Saliency in context},
  author={Jiang, Ming and Huang, Shengsheng and Duan, Juanyong and Zhao, Qi},
  booktitle={Proceedings of the IEEE conference on computer vision and pattern recognition},
  pages={1072--1080},
  year={2015}
}

@inproceedings{chen2020air,
  title={Air: Attention with reasoning capability},
  author={Chen, Shi and Jiang, Ming and Yang, Jinhui and Zhao, Qi},
  booktitle={European Conference on Computer Vision},
  pages={91--107},
  year={2020},
  organization={Springer}
}

@article{itti2002model,
  title={A model of saliency-based visual attention for rapid scene analysis},
  author={Itti, Laurent and Koch, Christof and Niebur, Ernst},
  journal={IEEE Transactions on pattern analysis and machine intelligence},
  volume={20},
  number={11},
  pages={1254--1259},
  year={2002},
  publisher={Ieee}
}

@article{itti2001computational,
  title={Computational modelling of visual attention},
  author={Itti, Laurent and Koch, Christof},
  journal={Nature reviews neuroscience},
  volume={2},
  number={3},
  pages={194--203},
  year={2001},
  publisher={Nature Publishing Group UK London}
}

@inproceedings{fan2018emotional,
  title={Emotional attention: A study of image sentiment and visual attention},
  author={Fan, Shaojing and Shen, Zhiqi and Jiang, Ming and Koenig, Bryan L and Xu, Juan and Kankanhalli, Mohan S and Zhao, Qi},
  booktitle={Proceedings of the IEEE Conference on computer vision and pattern recognition},
  pages={7521--7531},
  year={2018}
}

@inproceedings{jiang2014saliency,
  title={Saliency in crowd},
  author={Jiang, Ming and Xu, Juan and Zhao, Qi},
  booktitle={European conference on computer vision},
  pages={17--32},
  year={2014},
  organization={Springer}
}

@article{sun2024visual,
  title={How is visual attention influenced by text guidance? database and model},
  author={Sun, Yinan and Min, Xiongkuo and Duan, Huiyu and Zhai, Guangtao},
  journal={IEEE Transactions on Image Processing},
  volume={33},
  pages={5392--5407},
  year={2024},
  publisher={IEEE}
}

@inproceedings{mondal2023gazeformer,
  title={Gazeformer: Scalable, effective and fast prediction of goal-directed human attention},
  author={Mondal, Sounak and Yang, Zhibo and Ahn, Seoyoung and Samaras, Dimitris and Zelinsky, Gregory and Hoai, Minh},
  booktitle={Proceedings of the IEEE/CVF conference on computer vision and pattern recognition},
  pages={1441--1450},
  year={2023}
}

\end{document}